\pdfoutput=1
\documentclass{bmvc2k}

\title{A Tri-Layer Plugin to Improve \\ Occluded Detection}

\addauthor{Guanqi Zhan}{guanqi@robots.ox.ac.uk}{1}
\addauthor{Weidi Xie}{weidi@robots.ox.ac.uk}{1,2}
\addauthor{Andrew Zisserman}{az@robots.ox.ac.uk}{1}

\addinstitution{
 Visual Geometry Group\\
 University of Oxford\\
 Oxford, UK
}
\addinstitution{
Coop.\ Medianet Innovation Center \\ Shanghai Jiao Tong University \\ Shanghai, China
}

\runninghead{Zhan, Xie, Zisserman}{A Tri-Layer Plugin to Improve Occluded Detection}

\usepackage{bm}
\usepackage{bbding}
\usepackage{pifont}
\usepackage{amssymb}
\usepackage{comment}
\usepackage{multirow}
\usepackage{booktabs}
\usepackage{tabularx}
\usepackage{makecell}
\usepackage{color} \usepackage{appendix}

\usepackage{xcolor}
\usepackage{caption}
\definecolor{bmvc_blue}{RGB}{0,26,102} 
\begin{document}
\maketitle

\begin{abstract}
Detecting occluded objects still remains a challenge for state-of-the-art object detectors.
The objective of this work is to improve the detection for such objects, and thereby improve the overall performance of a modern object detector.

To this end we make the following four contributions:
(1) We propose a simple `plugin' module for the detection head of two-stage object detectors to improve the 
recall of partially occluded objects. The module predicts a tri-layer of  segmentation masks for the target
object, the occluder and the occludee, and by doing so is able to better predict the mask of the target object.
(2) We propose a scalable pipeline for generating training data for the module by using amodal completion of existing
object detection and instance segmentation training datasets to establish occlusion relationships.
(3) We also establish a COCO evaluation dataset to measure the recall performance of  partially occluded and separated objects.
(4) We show that the plugin module inserted into a two-stage detector can boost the performance significantly, by only fine-tuning the detection head, and with additional improvements if 
the entire architecture is fine-tuned. COCO results are reported for
Mask R-CNN with Swin-T or Swin-S backbones, and Cascade Mask R-CNN with a Swin-B backbone.
\end{abstract}

\section{Introduction}
\label{sec:intro}

\begin{figure*}[t]
	\centering
\includegraphics[height=0.18\linewidth]{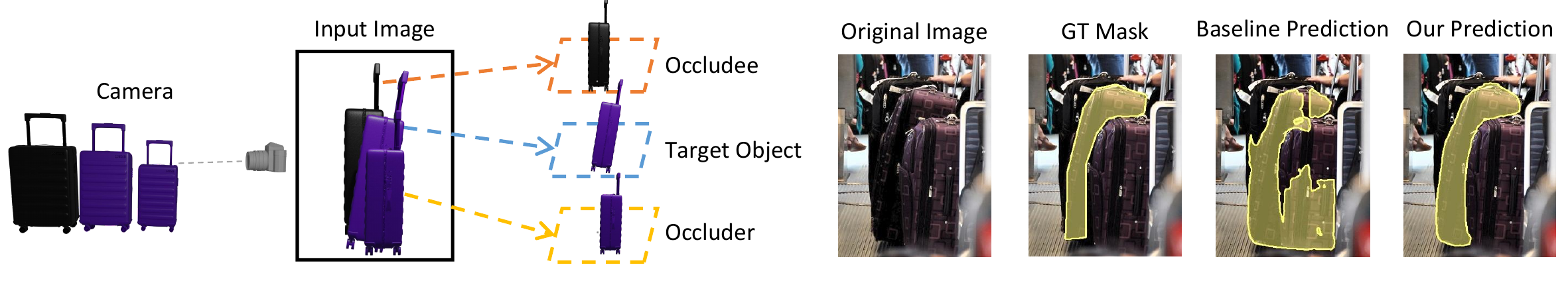}
	\caption{
	\textcolor{bmvc_blue}{\textbf{Improving occluded object detection and instance segmentation.} 
 Left: Occlusion is very common in the 3D world, where one object is in front of another and a portion of the scene disappears behind the non-transparent object that is closer to the viewer.
Right: For this example from COCO val,
a Swin-T + Mask R-CNN Baseline incorrectly detects and segments the target object (the middle of the three cases). However, if the detection head is replaced with our tri-layer plugin, a correct segmentation mask is obtained. The plugin is tasked with predicting the masks of the target object occluder and occludee, and this leads to a better modelling for occluded detection.
}
	} 
	\label{figure:teaser}
	\end{figure*}

Occlusion frequently occurs in images of real scenes and is both a benefit and a problem. It is a benefit, because it reveals depth orderings and so contributes to a 3D perception of the scene~\cite{Gibson79}. However, it is a problem because despite the continual increase in performance of object detectors over the last decade,
detection of occluded objects is still a
significant deficiency~\cite{wang2020robust,yuan2021robust,ke2021bcnet}.

In this paper, our objective is to improve the detection for objects under occlusion, and thereby improve the overall performance of the object detector. 
Specifically, we develop a lightweight `plugin' module, that can be inserted into the detection head of any two-stage object detector, 
{\em e.g.}~Mask R-CNN~\cite{mask_rcnn}, 
to improve the recall of occluded objects. 
The module simultaneously infers {\em three} segmentation `layers': the mask for the target object; the mask of the occluder~(the object in front that occludes the target); and the mask of the occludee~(the object behind that is occluded by the target) within the same detection box. This {\em tri-layer prediction head} forces the detection
 model to explicitly understand the existence of occlusion relationships, and thereby is better able to predict the target object mask. 
Additionally, we show that the process can be iterated, using the better prediction of the target mask to adapt the detection box for the next tri-layer prediction.

One challenge for training our proposed plugin module lies in the lack of suitable training data --
almost all large-scale detection datasets provide annotations on the visible part of objects, but no occlusion information is available.
To this end, we propose a scalable pipeline for automatically discovering  occluded objects (and their occluders), by running an amodal completion model on publicly available detection datasets, {\em e.g.}, COCO. 
Additionally, to verify the occlusion relationships between objects, 
{\em i.e.}, occluder or occludee,
we adopt an off-the-shelf monocular depth estimator~\cite{Ranftl2020} and determine the occlusion relationship based on their relative depth.
With this pipeline, we acquire a reliable large-scale data source for training our tri-layer model. We also
 establish an evaluation dataset to measure the recall performance for occluded objects, distinguishing between the two cases where the target object is {\em partially occluded} but the segmentation mask is connected, or where the target object segmentation mask is {\em separated} into distinct regions by the occluder.

To summarise, in this paper, we make the following four contributions:
(i) We propose a simple `plugin' module that can be inserted into two-stage object detectors, to improve their performance on occluded objects.
(ii) We establish a scalable pipeline to determine occlusion relationships between objects,
which is used to train the plugin module on publicly available detection datasets.
(iii) We set up an evaluation benchmark with real images to measure the recall performance of partially occluded and separated objects.
(iv) We show that the plugin module inserted into two-stage detectors can boost the performance significantly, by only fine-tuning the detection head, and with additional improvements if 
the entire architecture is fine-tuned. We show detection results for Mask R-CNN with Swin-T, Swin-S backbones, and Cascade Mask R-CNN with a Swin-B backbone.

\vspace{-0.4cm}
\section{Related Work}
\label{sec:related_work}

\paragraph{Object Detection \& Instance Segmentation.}
Methods for
object detection and instance segmentation have made great progress
by training deep neural networks on large-scale datasets~\cite{everingham2010pascal, coco_dataset, openimages_2020}.
Two-stage detectors like Faster R-CNN~\cite{faster_rcnn} and Mask R-CNN~\cite{mask_rcnn} first train region proposal networks to propose candidate objects. These candidates are then further refined in their location with a regression head, and categorised by a classifier head. With the development of stronger transformer-based~\cite{liu2021Swin, liu2021swinv2} architectures, replacing the original CNN backbones, there have been further improvements in the detection performance~\cite{liu2021Swin}. There has also been a similar development in single-stage detectors such as~\cite{carion2020detr, zhu2020deformable_detr, cheng2022masked} but we do not consider those in this work.
However, both single and two-stage detectors are still suffering when detecting occluded objects~\cite{saleh2021survey_occlusion}. 
\\[-0.9cm]

\paragraph{Amodal Segmentation \& Occlusion-Related Datasets.}
Amodal segmentation refers to the task of segmenting the object as whole, 
including the portions that are partially occluded~\cite{malik2016amodal, zhan2020self, sun2022amodal}. In the recent literature, 
various datasets have been collected to study occlusion~\cite{zhu2015cocoa, qi2019kins, chen2020transferable, cai2020messytable, wen2020ua}. However, most of these datasets  either focus on certain domains/tasks or do not provide a large number of images for 
training and evaluating models in terms of occlusion for a large variety of categories. 
\\[-0.9cm]

\paragraph{Layered Scene Understanding.}
Layered representation was originally proposed by Wang and Adelson~\cite{Wang94}
to represent a video as a composition of layers with simpler motions.
Since then, layered representations have been widely adopted in computer vision,
{\em e.g.}, to decompose videos into layers~\cite{brostow1999motion, Jojic01}, 
to improve scene segmentation by explicitly modelling occlusions, temporal consistency, depth ordering~\cite{winn2006layout, Kumar08,yang2010layered,yang2011layered},
to estimate optical flow~\cite{Sun12,Sun13,Wulff14,Wulff15}, and for novel view synthesis~\cite{zitnick2004high}. 
\\[-0.9cm]

\paragraph{Occlusion Handling.}
To improve object detection and instance segmentation under occlusion scenarios,
novel architectures, {\em e.g.},~compositional networks~\cite{wang2020robust, yuan2021robust}, 
have been proposed. Rather than develop a new architecture, ASN~\cite{qi2019kins} and ORCNN~\cite{follmann2019orcnn} ask modern detectors to output both modal and amodal masks of target objects. Additionally,
BCNet~\cite{ke2021bcnet} exploits two-layer graph neural networks 
to modern detectors for better instance segmentation under occlusion, inferring both the target object and the surrounding objects, 
while in this paper, we go one step further, and propose a simple, automatic pipeline for estimating objects' occlusion order, which enables training of the tri-layer plugin with explicit supervision on the occlusion ordering for objects. 

\vspace{-0.2cm}
\section{Detector Architecture and Application}
\vspace{-0.1cm}

In this section we describe our lightweight plugin module, and its application within an object detector. The module is designed to improve the detection performance on occluded objects.
We start by introducing the standard two-stage detector, and then describe the tri-layer plugin architecture and functionality.

\vspace{-0.4cm}
\paragraph{Two-stage detector.} 

Given an image detection dataset, $\mathcal{D} = \{(I_1, y_1), \dots, (I_n, y_n)\}$, 
a standard two-stage Mask R-CNN detector can be parametrized as:
\begin{align}
y_j = \{(b_j, c_j, m_j)\}^K = \Phi_{\{\textsc{CLS+BOX;SEG}\}} \circ \Phi_{\textsc{ALIGN}} \circ  \Phi_{\textsc{RPN}} \circ  \Phi_{\textsc{ENC}}(I_j)
\end{align}
where an input image~($I_j \in \mathbb{R}^{H \times W \times 3}$) with a total of $K$ objects is sequentially processed by a set of operations: 
an image encoder, $\Phi_{\textsc{ENC}}(\cdot)$; 
a region proposal network, $\Phi_{\textsc{RPN}}(\cdot)$; 
a region of interest feature alignment, $\Phi_{\textsc{ALIGN}}(\cdot)$;
after predicting the class and box offset~($\Phi_{\textsc{CLS + BOX}}(\cdot)$),
a binary mask for each RoI is also predicted, $\Phi_{\textsc{SEG}}(\cdot)$.
As a result, $y_j = \{(b_j, c_j, m_j)\}^K$ denotes the box coordinates~($b^k_j \in \mathbb{R}^4$),
category label~($c^k_j \in \mathbb{R}^{\mathcal{C}}$),
and \textbf{modal} segmentation mask~($m^k_j \in [0,1]^{H \times W}$) of the object,
which has been converted from the spatial resolution of RoI align to the original image.

In the following, we refer to a pair of objects that
have an occlusion relationship as  \textbf{occluder}~(the object that is in front of and thus occludes the other one), and  \textbf{occludee}~(the object at the back and being occluded), as illustrated in Figure~\ref{figure:teaser}.
Note that, the role of each object is often relative, 
thus it can be both occluder and occludee at the same time, depending on its paired objects. In Section~\ref{sec:generation_data}, we detail the procedure for acquiring the occlusion information,
including their estimated \textbf{amodal} segmentation masks and the occlusion orderings.

\begin{figure*}[t]
\label{sec:network_architecture}
	\centering
	\includegraphics[height=0.39\linewidth]{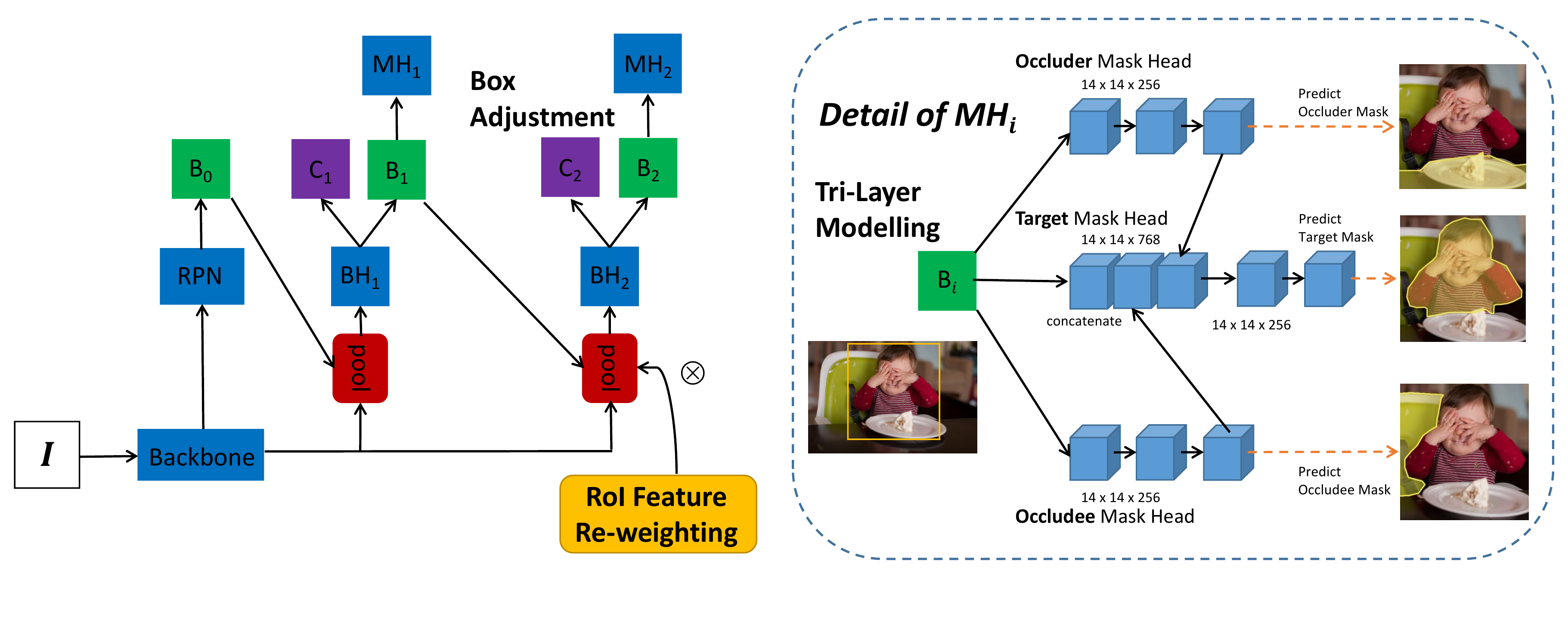}
	\vspace{-12pt}
\caption{
\textcolor{bmvc_blue}{\textbf{Architecture and function of the plugin module. } There are three functions: (a) The tri-layer mask head (MH$_i$ shown in detail on the right) predicts the mask of the target object (the infant wiping food on their face), the occluder (the dining table), and the occludee (the chair) within the detection box B$_i$.   The feature embeddings of the occluder/occludee branch are concatenated to the target mask embedding as cues to help better predict the target object mask; (b) As shown on the left, the process of predicting the target mask is iterated (index $i$), such that the second iteration is able to adjust the initial box predictions and better detect partially occluded / separated objects; (c) After the first iteration,
 RoI features are pooled according to the predicted target mask to guide the model to focus more on the partially occluded / separated object itself. The notation used is: ``$I$'' for the input image, and ``BH'', ``MH'', ``B'', ``C'' refer to bbox head, mask head, bounding box, and classification respectively.}
}
\label{figure:architecture}
\vspace{-0.6cm}
\end{figure*}

\vspace{-0.2cm}
\subsection{Architecture}
\vspace{-0.1cm}
In this section, 
we  introduce the architecture details of our tri-layer plugin for better detecting objects under occlusion. Specifically, in Section~\ref{sec:trilayer_modeling},
we augment the detection module to simultaneously output both occluder and occludee, 
along with the target object itself;
in Section~\ref{sec:adjust_box}, we introduce the idea of box adjustment, 
that facilitates the model to reason about the full object coverage with only partial observation, 
due to partial occlusion or separation; and
in Section~\ref{sec:weighted_RoI_pooling}, 
we improve the RoI pooling procedure by weighting the feature map with the inferred object mask, effectively preventing the model from concentrating on the occluder/occludee.
Figure~\ref{figure:architecture} illustrates the architecture of the plugin.

\vspace{-0.2cm}
\subsubsection{Tri-Layer Modelling}
\vspace{-0.1cm}
\label{sec:trilayer_modeling}

Here, we augment the instance segmentation head with a tri-layer module that accepts the RoI aligned feature map as input, 
and outputs the modal segmentation masks of the occluder and occludee for the target object respectively~(denoted as \textbf{Occluder Mask Head} and \textbf{Occludee Mask Head}). 
Specifically, we pass the RoI aligned feature maps into three different mask prediction heads: 
\begin{align}
    \{\hat{b}_j, \hat{c}_j, \hat{m}_j, \hat{m}_{j1}, \hat{m}_{j2}\}^K = \Phi_{\{\textsc{CLS+BOX;SEG}\}}(\mathcal{F}_j)
\end{align}
where $\mathcal{F}_j$ refers to the feature map from RoI align, and
$\hat{m}_{j1}, \hat{m}_{j2}$ refer to the inferred modal segmentation masks for the object's occluder and occludee respectively. 
Note that, an object may not have occluder or occludee, and the predicted mask in this case is all zero.

As shown in Figure~\ref{figure:architecture}~(right), 
the feature embeddings from occluder/occludee branches are further integrated into the target object segmentation branch~($\hat{m}_j$), providing cues to better infer the modal mask of the target object. 
To act as a proper layering model, we distinguish the order of occluder and occludee by concatenating the feature embeddings in order. Note that, this is in contrast to the previous approach~\cite{ke2021bcnet},
where the embeddings are simply element-wise added, 
leading to ambiguous occlusion ordering from the commutative rule.

\vspace{-0.2cm}
\subsubsection{Box Adjustment}
\label{sec:adjust_box}
\vspace{-0.1cm}

After predicting the box and segmentation, 
we carry out a second iteration, to refine the initial predictions. 
Intuitively, if the network can detect part of an occluded object in the first iteration, 
an extra iteration will provide the opportunity to adjust the box accordingly, 
to include any part of the occluded object missed in the first iteration. For the $k$th instance in image $j$,
\begin{align}
   \{\dot{b}_j, \dot{c}_j, \dot{m}_j, \dot{m}_{j1}, \dot{m}_{j2}\}^k = \Phi_{\{\textsc{CLS+BOX;SEG}\}} (\Phi_{\textsc{ALIGN}}(\mathcal{V}_j, \hat{b}_j^k)), \hspace{3pt} \forall k \in [1,K]
\end{align}
where $\mathcal{V}_j$ denotes the feature map from the visual encoder. 
The inferred object box~($\hat{b}_j^k$) from the first iteration is used for RoI align to generate refined boxes~($\dot{b}_j^k$), and then we can do tri-layer modelling in the refined boxes.

\vspace{-0.2cm}
\subsubsection{RoI Feature Re-weighting} 
\label{sec:weighted_RoI_pooling}
\vspace{-0.1cm}

To guide the model to focus more on the target objects, 
rather than the occluders/occludees that may take up a large proportion of the box, 
we also introduce RoI feature re-weighting with the inferred object mask from the previous iteration:
\begin{align}
   \{\dot{b}_j, \dot{c}_j, \dot{m}_j, \dot{m}_{j1}, \dot{m}_{j2}\}^k = \Phi_{\{\textsc{CLS+BOX;SEG}\}}  (\Phi_{\textsc{ALIGN}}(\mathcal{V}_j, \hat{b}_j^k) \otimes \hat{m}_j^k), \hspace{3pt} \forall k \in [1,K]
\end{align}
where $\hat{m}_j^k$ denotes the inferred segmentation mask from the previous iteration, 
$\otimes$ denotes the element-wise product between the RoI aligned feature map and the object segmentation from the previous iteration, and $\dot{b}_j, \dot{c}_j, \dot{m}_j$ refer to the output bounding box, 
object category and instance mask, respectively.

\vspace{-0.2cm}
\subsubsection{Training} 
\vspace{-0.1cm}
\label{sec:method_training}

We start from publicly released pre-trained models~\cite{swin_object_detection_github}. The fine-tuning procedure is conducted progressively from ``Tri-Layer Modelling'' to ``BBox Adjustment'' and then ``RoI Feature Re-weighting'', {\em i.e.},~proposed modules are gradually added after the previous one converges. 
For tri-layer modelling, both occluder and occludee mask heads use the same structures as the main mask head, except that they are class-agnostic while the main mask head is not. They are trained with binary cross-entropy loss using the inferred ground truth masks, as will be detailed in Section~\ref{sec:occlusion_reasoning}. 
The implementation and training are based on MMDet~\cite{mmdetection}.
Please refer to the appendix for more training details.

\vspace{-0.2cm}
\section{Data Preparation}
\label{sec:generation_data}
\vspace{-0.1cm}

To properly train our plugin,
ground truth occluder/occludee masks for each object are required,
as discussed in Section~\ref{sec:method_training}.
In this section, we describe the procedure for determining objects' occlusion order based on their \textbf{amodal} segmentation and depth ordering.

In general, acquiring amodal segmentations can be very costly, 
due to the requirement of annotating the occluded parts of the object.
To our knowledge, none of the existing large-scale datasets, 
{\em e.g.}~COCO~\cite{coco_dataset}, LVIS~\cite{gupta2019lvis}, provide amodal segmentation masks. Here, we start by describing a simple, automatic, 
thus scalable pipeline to approximate object amodal segmentation masks~(Section~\ref{sec:amodal_completion});
in Section~\ref{sec:occlusion_reasoning}, 
we further exploit these amodal segmentations to infer the occlusion relationships between paired objects; and in Section~\ref{sec:occluded_separated}, 
we detail two different types of occlusion, namely, separation and partial occlusion,
and define an evaluation benchmark for occlusion, based on the COCO2017 val set. 
\vspace{-0.2cm}
\subsection{Amodal Completion}
\label{sec:amodal_completion}
\vspace{-0.1cm}

Amodal completion aims to infer the amodal mask for an object, given its modal mask, and we want to do amodal completion on COCO.
Here, we adopt a similar approach to that of~\cite{zhan2020self}, but re-train the amodal completion model on COCO to reduce the domain gap. 
In detail, instance masks are selected from the COCO dataset and randomly pasted onto  training images in order to create artificial occlusions in the images. The amodal completion
model is then trained on these images with artificial occlusions to predict the occluded parts, 
conditioned on the partial (modal) observation.
During inference time, 
for each object, we slightly dilate its modal mask to reduce the gap between connected COCO masks, and the model can thus infer the object's mask to its full extent, {\em i.e.}, amodal mask.

While evaluating for amodal completion on different datasets, 
as shown in Table~\ref{table:amodal_completion}, 
our model brings a significant improvement over previous approach on COCO.
We refer the readers to appendix for more evaluation details.

\captionsetup{margin=12pt , font=small,  labelfont={color=bmvc_blue,bf}, labelsep=period, skip=5pt}
\hspace{-0.3cm}
\begin{minipage}[t]{0.5\textwidth}
\setlength{\tabcolsep}{8pt}
\centering
\footnotesize
\begin{tabular}{c|c|c}
Model & Eval Dataset & mIoU   \\ \toprule 
~\cite{zhan2020self} & COCOA val &	81.35  \\ 
~\cite{zhan2020self} & COCO2017 val &	69.32  \\ 
Ours & COCO2017 val &	\textbf{81.55} \\ \bottomrule
\end{tabular}
\captionof{table}{\textcolor{bmvc_blue}{\textbf{Amodal Completion on COCO:}
The comparison between our model and~\cite{zhan2020self}}}
\label{table:amodal_completion}
\end{minipage}
\hspace{-0.6cm}
\begin{minipage}[t]{0.5\textwidth}
\setlength{\tabcolsep}{8pt}
\footnotesize
\centering
\begin{tabular}{c|c}
Dataset & \# Total Objects   \\ \toprule 
Separated COCO  &  3522  \\ 
Occluded COCO   &  5550  \\ 
Occluder Masks  &  345169  \\ 
Occludee Masks  &  328561  \\ \bottomrule
\end{tabular}
\captionof{table}{\textcolor{bmvc_blue}{\textbf{Statistics of our generated training and evaluation datasets.}}}
\label{table:statistics_dataset}
\end{minipage}

\vspace{-0.4cm}
\subsection{Occlusion Reasoning}
\label{sec:occlusion_reasoning}
\vspace{-0.1cm}

\captionsetup{margin=0pt , font=small,  labelfont={color=bmvc_blue,bf}, labelsep=period, skip=5pt}
\begin{figure*}[t]
		\centering
		\includegraphics[trim=0.5cm 0cm 0.5cm 0cm, height=0.21\linewidth]{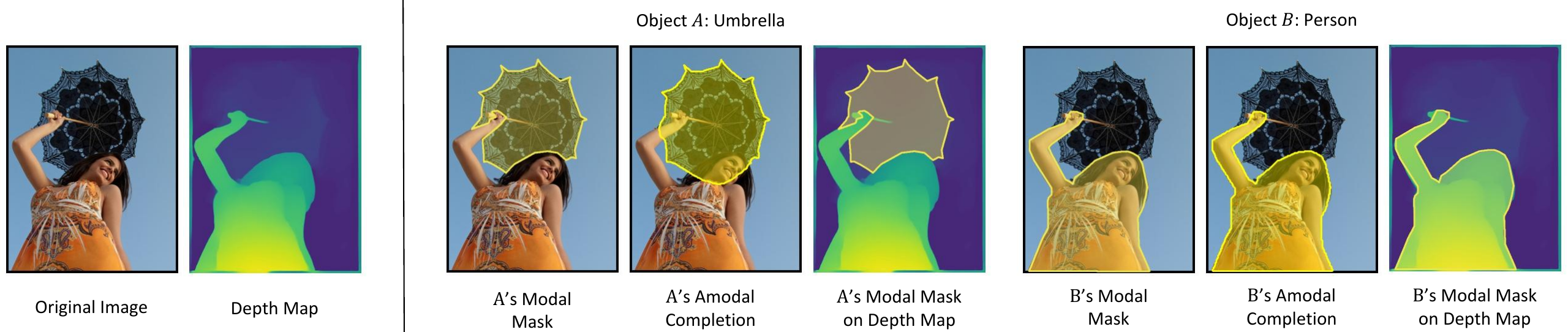}
		\caption{\textcolor{bmvc_blue}{ \textbf{The process of occlusion reasoning via amodal completion and depth estimation.} Left: The original image and its depth map. Right: the amodal completion and depth maps for the two objects: Umbrella and Person. In this case we conclude that the Person occludes the Umbrella since: (i) the amodal umbrella has an overlap with the modal person, but no overlap vice versa; and (ii) the average depth map indicates that the depth of the umbrella is greater than that of the person. In this way, we predict that \textbf{``The umbrella is occluded by the person''.} }
  }
		\label{figure:occlusion_reasoning}
 		\vspace{-6mm}
\end{figure*}

Once we get the amodal mask for each object, 
the occlusion ordering between a pair of connected objects is inferred in two stages (most easily understood by first looking at the example in Figure~\ref{figure:occlusion_reasoning}):
{\em First}, we compute the intersection between one object's amodal segmentation and the other object's modal segmentation. Specifically, for each pair of connected objects $A$ and $B$,
the intersection of $A$'s amodal mask with $B$'s modal mask can be denoted as $I_A$, 
and the intersection of $B$'s amodal mask with $A$'s modal is denoted as $I_B$.
Then $A$ is likely to be occluded by $B$, if $I_A > I_B$.
{\em Second}, we verify the results by depth estimation, 
{\em i.e.},~the occludee should have a greater depth than the occluder. 
We adopt an off-the-shelf depth estimator~\cite{Ranftl2021}. After conducting inference on all COCO images, we compute the average depth of each object over all pixels. Denoting object $A$'s average depth as $d_A$, $B$'s average depth as $d_B$. 
If $d_A > d_B$ also holds, the result of occlusion ordering is then verified. 
In this way, we determine the occlusion relationship between objects with greater confidence.
For those cases with inconsistent amodal completion and depth verification, 
we do not assign any occluder-occludee relationship.

With such occlusion reasoning, for each confirmed object pair
we have the pseudo ground truth relative occlusion order and ground truth modal masks, 
to train the tri-layer module~(described in Section~\ref{sec:trilayer_modeling}).
Note that, when there are multiple occluders/occludees for one target object,
we merge all their modal masks to form the final occluder/occludee mask.

\begin{figure*}[t]
		\centering
		\includegraphics[trim=0.5cm 0cm 0.5cm 0cm, height=0.25\linewidth]{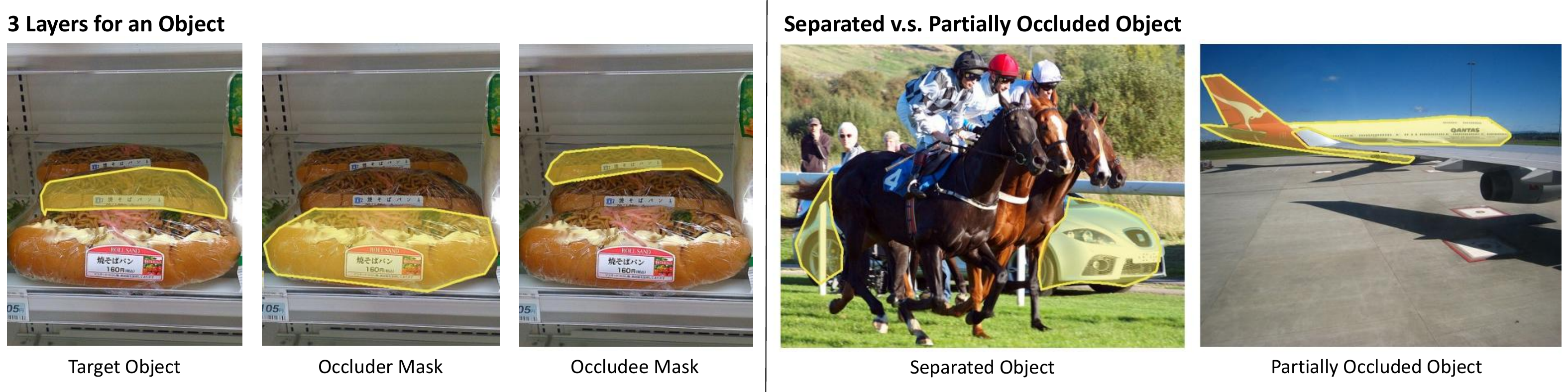}
		\caption{\textcolor{bmvc_blue}{\textbf{Examples in generated training and testing datasets.} Left: Example of a target object with its occluder and occludee. Right: Examples of automatically picked objects for Separated COCO and Occluded COCO. More examples are in appendix.}}
		\label{figure:separated_occluded}
 		\vspace{-4mm}
\end{figure*}

\vspace{-0.2cm}
\subsection{Occluded COCO \& Separated COCO for Evaluation}
\label{sec:occluded_separated}
\vspace{-0.1cm}

To monitor progress of object detection and instance segmentation under occlusion, 
we create a benchmark based on the COCO2017 val dataset.
We start by defining two different splits for occluded objects, 
namely, \textbf{Separated COCO} and \textbf{Occluded COCO}, consisting of \textbf{separated} and \textbf{partially occluded} objects respectively.
Unlike previous work~\cite{wang2020robust} that manually selects occluded objects in COCO, 
we collect the data splits automatically, based on the occlusion reasoning.
Specifically, for each object, we can easily check its ground truth modal mask for connectivity. 
If the mask is split into pieces, then the object is ``seperated'' and should be in the Separated COCO split,
for example, the car behind the horses in Figure~\ref{figure:separated_occluded};
otherwise, the object is put into the Occluded COCO split as ``partially occluded'' object, if it is confirmed to be occluded by some other objects, 
{\em e.g.},~the airplane at the back in the last image of Figure~\ref{figure:separated_occluded}.
In this way, we can easily collect two datasets with different occlusion types.
Detailed statistics are given in Table~\ref{table:statistics_dataset}.

\vspace{-0.2cm}
\section{Experiments}
\vspace{-0.1cm}

\subsection{Datasets and Implementation Details}
\label{sec:implementation_details}
\vspace{-0.1cm}

\paragraph{Datasets:} 
We train all models on the COCO2017 training split, 
together with the occluder/occludee masks obtained in the way described in Section~\ref{sec:occlusion_reasoning}.
We evaluate the performance to detect occluded objects on Separated COCO and Occluded COCO generated in Section~\ref{sec:occluded_separated}, and the overall detection performance on both COCO2017 val and COCO2017 test-dev. 
As for evaluating generalisability, we also test on other datasets like KINS~\cite{qi2019kins} (7517 images for testing), OVIS~\cite{qi2022ovis} and OpenImages~\cite{openimages_2020} (details in appendix). \\[-0.8cm]

\paragraph{Baselines:} 
We compare the following state-of-the-art architectures,
 Swin-T + Mask R-CNN, Swin-S + Mask R-CNN and Swin-B + Cascade Mask R-CNN, with or without our designed plugin. 
In addition, we also compare our tri-layer plugin with the bi-layer modelling approach~(inspired by BCNet~\cite{ke2021bcnet}, set all connected objects to be occluders), and compositional network~\cite{yuan2021robust} that is specifically designed to handle object occlusion with instance masks. \\[-0.8cm]

\paragraph{Evaluation Metrics:} 
In addition to the standard metric for COCO2017 val and COCO2017 test-dev,
{\em i.e.},~mAP, we also calculate the recall on Occluded COCO and Separated COCO, 
\emph{i.e.},~the number of partially occluded / separated objects that are recalled with a fixed number of RPN proposals. 
Specifically, we treat a partially occluded / separated object being recalled if and only if there is a detection whose $\text{confidence} > 0.3$ and mask $\text{IOU} > 0.75$ with it. 
mIOU on KINS is calculated, in accordance with that in~\cite{yuan2021robust}.\\[-0.6cm]

\subsection{Ablation Study}
\label{sec:ablation_study}
\vspace{-0.1cm}

Here, we conduct thorough experiments to validate the effectiveness of different modules in our plugin. As shown in Table~\ref{table:ablation_study}, we make the following observations: 
1) \textbf{Tri-layer modelling}:
brings a significant improvement in terms of Recall on Occluded COCO (B1 to B3, +75, when only fine-tuning the mask heads; B1 to C2, +96, when fine-tuning the whole network);
2) \textbf{Box adjustment}: gives a significant performance boost for BBox mAP, for example, from B1 to B2,  +1.9 for only fine-tuning the head, and from B1 to C1, +2.3 for fine-tuning the network;
3) \textbf{RoI feature re-weighting}: further improves  Mask mAP, Recall on Occluded COCO and Recall on Separated COCO (+0.3/+10/+21 for only fine-tuning the head and +0.1/+7/+15 for fine-tuning the whole network), which is shown by the models, from B4 to B5, and C3 to C4;
4) \textbf{Fine-tuning the whole network}: 
can generally bring improvement on all evaluation metrics,
for example, comparing B2-B5 with C1-C4. Notably, only fine-tuning the head could already contribute the majority of the improvement, validating the effectiveness of our proposed module as a general `plugin', which can be inserted into pre-trained detectors, and give quick performance improvement.

\begin{table}[h]
\setlength{\tabcolsep}{6pt}
\footnotesize
\centering
\tabcolsep=0.08cm
\begin{tabular}{c|cccc|cccc}
\hline
  Model &	\thead{Tri-Layer \\ Modelling}	& \thead{BBox \\ Adjustment} &	\thead{RoI Feature \\ Re-weighting} & \thead{Fine-tuning \\ Whole Network?} & \thead{Recall \\ Occluded} & \thead{Recall \\ Separated} & \thead{BBox \\ mAP} & \thead{Mask \\ mAP}   \\ \hline 
B1 & & & &  & 3264(58.81\%) & 1125(31.94\%) & 46.0 & 41.6	 \\
B2 &  & \checkmark & &  & 3296(59.39\%) & 1141(32.40\%) & 47.9 & 42.2  \\
B3 & \checkmark &  & &  & 3339(60.16\%) & 1157(32.85\%) & 46.0 & 41.9  \\
B4 & \checkmark & \checkmark & &  & 3400(61.26\%) & 1187(33.70\%) & 48.1 & 42.5 \\
B5 & \checkmark & \checkmark & \checkmark &  & \textbf{3410(61.44\%)} & \textbf{1208(34.30\%)} & \textbf{48.2} & \textbf{42.8} \\
\hline
C1 &  & \checkmark & &  \checkmark & 3367(60.67\%) & 1170(33.22\%) & 48.3 & 42.5 \\ 
C2 & \checkmark & & &  \checkmark & 3360(60.54\%) & 1159(32.91\%) & 46.3 & 42.2 \\
C3 & \checkmark & \checkmark & &  \checkmark & 3434(61.87\%) & 1208(34.30\%) & 48.3 & 42.9 \\  
C4 & \checkmark & \checkmark & \checkmark &  \checkmark & \textbf{3441(62.00\%)} & \textbf{1223(34.72\%)} & \textbf{48.5} & \textbf{43.0} \\ \hline
\end{tabular}
\caption{\textcolor{bmvc_blue}{\textbf{Ablation study} for adding our plugin to Swin-T + Mask R-CNN. Note that B1-B2 only mask heads are fine-tuned while B2-B3, B3-B4 only bbox heads and mask heads are fine-tuned.} 
}
\vspace{-0.3cm}
\label{table:ablation_study}
\end{table}

\subsection{Comparison with State-of-the-Art}
\label{compare_stoa}

\paragraph{Comparison on COCO.} 
As shown in Table~\ref{table:compare_stoa},
we inject our plugin into a series of popular strong architectures, 
{\em i.e.}~Mask R-CNN / Cascade Mask R-CNN with different Swin Transformers as backbone. In all cases, the plugin can always improve recalls for both Occluded COCO and Separated COCO, as well as detection performance on BBox and Mask mAP, 
sometimes by over 2.5/1.4 (val) and 2.4/1.4 (test-dev) mAP on box and mask predictions. In particular, 
when compared with bi-layer modelling
,
our plugin on Swin-T + Mask R-CNN can recall 126 more objects on Occluded COCO and 76 more on Separated COCO, and boost BBox/Mask mAP by 2.2/1.0 (val) and 2.2/1.1 (test-dev) respectively,
which shows the effectiveness of our plugin.

\begin{table}[!htb]
\setlength{\tabcolsep}{10pt}
\footnotesize
\centering
\tabcolsep=0.10cm
\begin{tabular}{ccccccccc}
\toprule
\multirow{2}*{Detector} & \multirow{2}*{Backbone} & \multirow{2}*{Plugin} & 
\multirow{2}*{\thead{Recall \\ Occluded}} &
\multirow{2}*{\thead{Recall \\ Separated}} &
\multicolumn{2}{c}{val mAP} & \multicolumn{2}{c}{test-dev mAP} \\
\cmidrule(lr){6-7}\cmidrule(lr){8-9}
& & & & & \thead{BBox} & \thead{Mask} & \thead{BBox} & \thead{Mask} \\
  \midrule
Mask R-CNN &	Swin-T~\cite{swin_object_detection_github} & -- & 3264(58.81\%) & 1125(31.94\%) & 46.0 & 41.6 & 46.3 & 42.0 \\ 
Mask R-CNN &	Swin-T & bi-layer & 3315(59.73\%) &	1147(32.57\%) & 46.3 & 42.0 & 46.5 & 42.3 \\ 
Mask R-CNN & Swin-T & ours & \textbf{3441(62.00\%)} & \textbf{1223(34.72\%)} & \textbf{48.5} & \textbf{43.0} & \textbf{48.7} & \textbf{43.4} \\ \midrule
Mask R-CNN & Swin-S~\cite{swin_object_detection_github} & -- & 3393(61.14\%) & 1186(33.67\%) & 48.5 & 43.3 & 49.0 & 44.1 \\ 
Mask R-CNN & Swin-S & ours & \textbf{3473(62.58\%)} & \textbf{1261(35.80\%)} & \textbf{50.3} & \textbf{44.2} & \textbf{50.6} & \textbf{44.9} \\ \midrule
Cascade Mask R-CNN & Swin-B~\cite{swin_object_detection_github} & -- & 3491(62.90\%) & 1279(36.31\%) & 51.9 & 45.0 & 52.6 & 45.6 \\ 
Cascade Mask R-CNN & Swin-B & ours* & \textbf{3532(63.64\%)} & \textbf{1299(36.88\%)} & \textbf{52.1} & \textbf{45.4} & \textbf{52.7} & \textbf{45.9} \\ \bottomrule
\end{tabular}
\caption{\textcolor{bmvc_blue}{
\textbf{Comparison with state-of-the-art on different architectures.} 
The plugin gives a  performance boost across 
all the architectures, 
even for the strongest detector~(Swin-B + Cascade Mask R-CNN).
* Only Tri-Layer Modelling is applied as Cascade Mask R-CNN has already used multiple iterations. 
}}
\vspace{-10pt}
\label{table:compare_stoa}
\end{table}

\vspace{-15pt}
\paragraph{Comparison on other benchmarks.}
Here, we directly evaluate the model on the KINS~\cite{qi2019kins} dataset in terms of mIoU. 
To handle the problem that KINS classes and COCO classes are different, we make a mapping from COCO classes to KINS classes, as detailed in the appendix.
Note that, these models are only trained on COCO, 
thus resembling a cross-domain generalisation.
Specifically, 
while evaluating the baseline Swin-T + Mask R-CNN, it only achieves 66.6 mIoU,
which is slightly lower than the CompositionalNet instance segmentation work~\cite{yuan2021robust} with a 67.2 mIoU. However, with our plugin module, the performance can be largely improved, 
getting 68.5 mIoU and becoming better than~\cite{yuan2021robust} on KINS dataset.
We refer the reader to the appendix for more detailed results on KINS, OVIS, and OpenImages.

\begin{table}[h]
\setlength{\tabcolsep}{6pt}
\footnotesize
\centering
\tabcolsep=0.08cm
\begin{tabular}{ccccccc}
\hline
  Detector & Backbone &	\thead{Tri-Layer \\ Modelling}	& \thead{BBox \\ Adjustment} &	\thead{RoI Feature \\ Re-weighting} & \# Parameters & FLOPs  \\ \hline 
Mask R-CNN & Swin-T & & &  & 47.8M & 263.78G	 \\
Mask R-CNN & Swin-T & \checkmark & &  & 54.2M & 	389.87G \\
Mask R-CNN & Swin-T & \checkmark & \checkmark &  & 77.6M & 583.33G	 \\
Mask R-CNN & Swin-T & \checkmark & \checkmark & \checkmark & 77.6M & 583.33G	 \\ \hline 
Mask R-CNN & Swin-S & & &  & 69.1M & 353.77G	 \\
Mask R-CNN & Swin-S & \checkmark & &  & 75.5M & 	479.86G \\
Mask R-CNN & Swin-S & \checkmark & \checkmark &  & 98.9M & 673.32G	 \\
Mask R-CNN & Swin-S & \checkmark & \checkmark & \checkmark & 98.9M & 673.32G	 \\ \hline 
Cascade Mask R-CNN & Swin-B & & &  & 145.0M & 975.44G	 \\
Cascade Mask R-CNN & Swin-B & \checkmark & &  & 164.3M &	1353.68G \\ \hline 

\end{tabular}
\caption{\textcolor{bmvc_blue}{
\textbf{Comparison of number of parameters and FLOPs.} ``Tri-Layer Modelling''
introduces two extra mask heads, but only increases the number of parameters by a small proportion. The parameter increase brought by ``BBox Adjustment'' is mainly due to the extra bbox head (14M) in the extra iteration. The increase in FLOPs is approximately proportional to the increase in the number of parameters. }
}
\vspace{-0.7cm}
\label{table:num_parameter}
\end{table}

\paragraph{Comparison of number of parameters and FLOPs.}
Table~\ref{table:num_parameter} compares the number of parameters and FLOPs for different models with/without our plugin. ``Tri-Layer Modelling'' is a lightweight plugin, only introducing 9.3\%, 13.4\%, 13.3\% more parameters for Swin-S + Mask R-CNN, Swin-T + Mask R-CNN, and Swin-B + Cascade Mask R-CNN, respectively.
``BBox Adjustment'' introduces more parameters for each model, where the majority of increase comes from the extra bbox head (14M) in the extra iteration. ``RoI Feature Re-weighting'' does not introduce any extra parameters since it only re-weights the RoI feature using the inferred segmentation mask from the previous iteration. When only fine-tuning the heads, only a small proportion of parameters need to be adjusted, so the training speed is fast.

\subsection{Qualitative Results}
\label{qualitative_results}

\begin{figure*}[!htb]
		\centering
		\includegraphics[trim=0.5cm 0cm 0.5cm 0cm, height=0.43\linewidth]{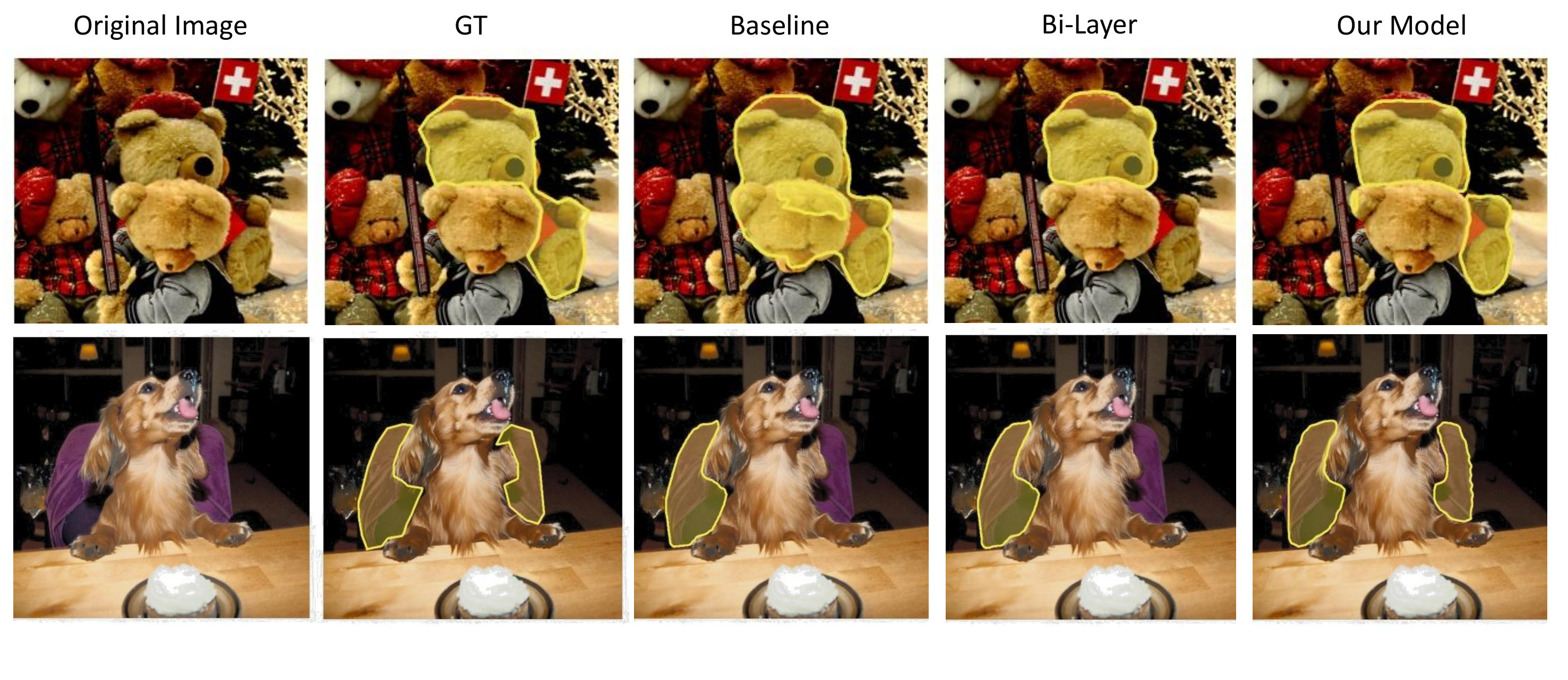}
		\vspace{-6mm}
		\caption{\textcolor{bmvc_blue}{\textbf{Qualitative results on COCO.} 
		Please see the text for more discussion. More qualitative results are provided in appendix.
  }}
		\label{figure:qualitative_compare}
 		\vspace{-6mm}
\end{figure*}

In Figure~\ref{figure:qualitative_compare}, 
we show qualitative results for inserting our plugin into Swin-T + Mask R-CNN.
As can be seen, the baseline model tends to fail in challenging occlusion cases, 
either over-segmenting the partially occluded~(row 1) or under-segmenting the separated~(row 2) objects.
While our proposed model has largely improved the detection, 
for example, disambiguating the teddy bears~(row 1), 
and inferring the two separated pieces of the chair that is heavily occluded by the dog~(row 2). See appendix for more examples.\\[-0.7cm]

\vspace{-0.3cm}
\section{Conclusion and Future Work}
\vspace{-0.1cm}

We have proposed a simple `plugin' module for two-stage object detectors that can improve their performance in detecting objects under challenging occlusions. 
Additionally, we describe a scalable pipeline for automatically identifying occluded and occluding objects in existing benchmarks, to provide training data and an evaluation dataset. Adding the module to a series of popular strong detectors, Mask R-CNN / Cascade Mask R-CNN with different Swin Transformer backbones, leads to consistent performance improvements.

A possible avenue of future work is to improve detection performance of occluded objects in videos, where multiple views of the objects and temporal cues are potentially  available to help disambiguate the occlusions.

\paragraph{Acknowledgements. } 
This research is supported by EPSRC Programme Grant VisualAI EP$\slash$T028572$\slash$1, a Royal Society
Research Professorship RP$\backslash$R1$\backslash$191132, and a China Oxford Scholarship. 
We thank Prannay Kaul, Tengda Han and Gyungin Shin for proof-reading.

\bibliography{egbib}
\clearpage

\appendix
\vspace{6mm}
\renewcommand{\appendixpagename}{\color{bmvc_blue} Appendix}
\appendixpage
\addappheadtotoc

\vspace{6mm}


\section{Training Details}
\label{sec:training_details}
Our model is implemented with MMDet
~\cite{mmdetection}, an open-source object detection toolbox based on Pytorch. For each fine-tuning process, it takes about 20 epochs to converge. We set the initial learning rate to be 0.000001, which is the learning rate at the end of the official COCO training of these models~\cite{swin_object_detection_github}, and drop the learning rate by 10 at epoch 16. Weight decay is set to be 0.05, and batch size is 16, following the official training setting. When only the head is fine-tuned, the training process is much quicker because the number of trained weights is greatly cut down. Approximately the training time will be only half of that for fine-tuning the entire network.

\section{Amodal Completion}
\label{sec:sup_amodal_completion}
In Section~\ref{sec:sup_amodal_quantitative}, we provide details about the evaluation of the amodal completion model. 
After that, in Section~\ref{sec:sup_amodal_qualitative_comparison}, we show more visualisation examples to illustrate the effectiveness of the amodal completion model on COCO val.

\subsection{Details of Quantitative Evaluation}
\label{sec:sup_amodal_quantitative}

In this section, we aim to evaluate the performance of different models for amodal completion on COCO. 
However, one challenge is that COCO does not provide GT amodal masks.
We thus borrow the GT amodal masks from COCOA~\cite{zhu2015cocoa} which is a subset of COCO with manual annotation of the amodal segmentation mask for each object. 
We transfer the GT amodal masks from COCOA to COCO as follows: first, determine the images that are in common between the two datasets; then for a specific object in COCO, within a common image,  determine if COCOA provides an amodal mask by comparing their modal masks using IoU. This is necessary because the modal mask in COCOA might be slightly different from that in COCO.  If the IoU is greater than 0.7, then the match is accepted, and the amodal mask is used.
As a result, we evaluate on 450 objects in COCO2017 val.

As Table~\ref{table:amodal_completion} shows (result of ~\cite{zhan2020self} on COCOA val is as reported in their paper), the performance of ~\cite{zhan2020self} drops significantly when applied to COCO2017 val. In contrast, our model achieves better performance on COCO2017 val than~\cite{zhan2020self} on COCOA val. 
In Section~\ref{sec:sup_amodal_qualitative_comparison}, 
we provide qualitative results from both our model and~\cite{zhan2020self}.
We conjecture that the substantial performance drop of~\cite{zhan2020self} is due to the domain gap between COCO and COCOA, {\em e.g.}, the annotations in COCO tend to have a gap between objects, and even for the same object the modal annotation mask in COCO and COCOA might be slightly different, thus the model trained on COCOA will struggle to generalise to COCO val.

\subsection{Qualitative Comparison}
\label{sec:sup_amodal_qualitative_comparison}
\begin{figure*}[!htb]
		\centering
		\includegraphics[trim=0.5cm 0cm 0.5cm 0cm, width=0.8\linewidth]{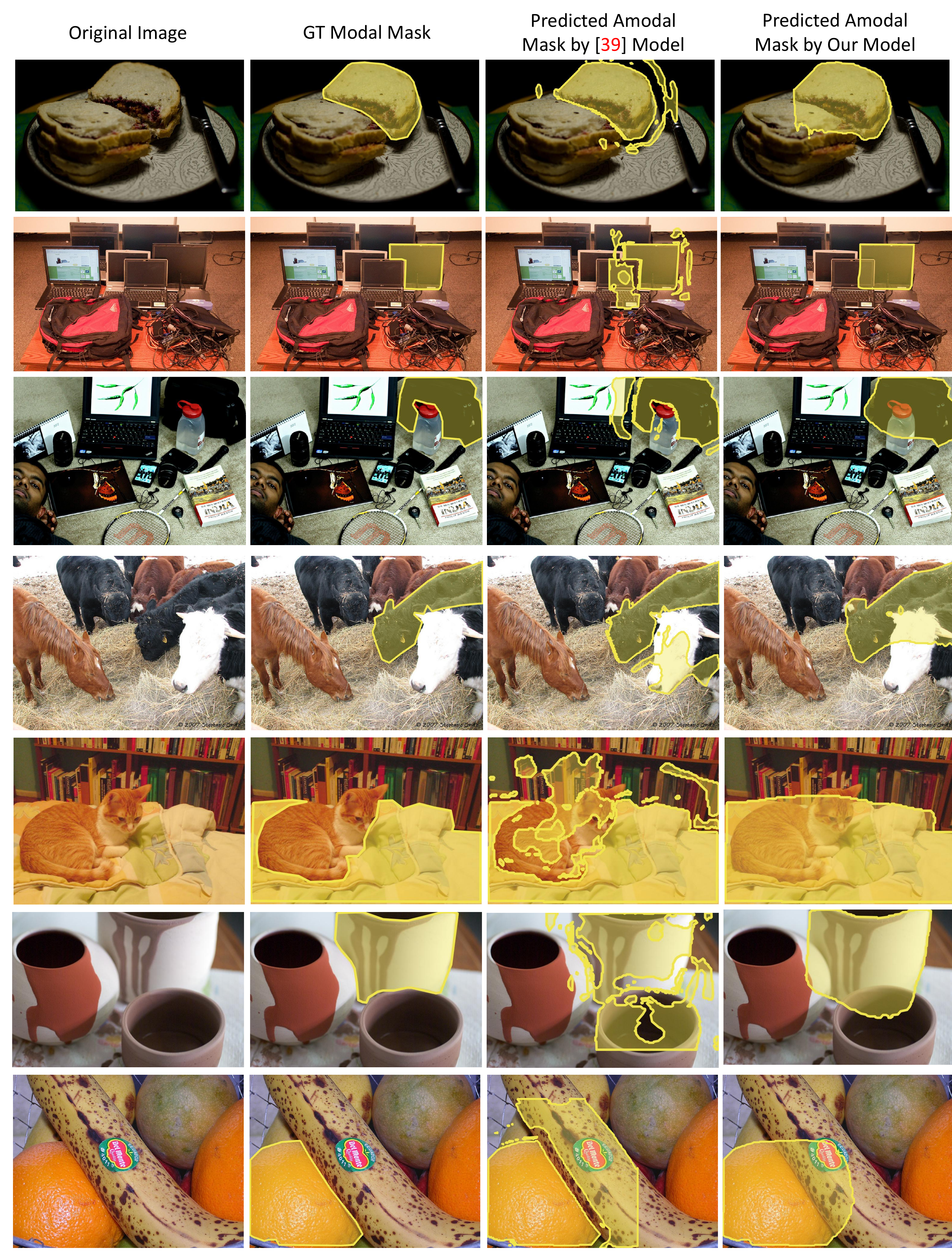}
		\vspace{-2mm}
		\caption{\color{bmvc_blue} \textbf{Comparison of different amodal completion models.} Our amodal completion model performs significantly better than the model of~\cite{zhan2020self}.}
		\label{figure:sup_amodal_completion}
 		\vspace{-4mm}
\end{figure*}

Figure~\ref{figure:sup_amodal_completion} shows qualitative results of our model and the model of~\cite{zhan2020self}
for amodal completion on COCO2017 val. 
We can observe that our model can generate significantly better amodal segmentation masks, and reason about the occluded parts of the objects.

\clearpage

\section{Examples of Generated Training Data}
\label{sec:sup_generated_training_data}
\begin{figure*}[h!]
		\centering
		\includegraphics[trim=0.5cm 0cm 0.5cm 0cm, 
  width=0.8\linewidth
  ]{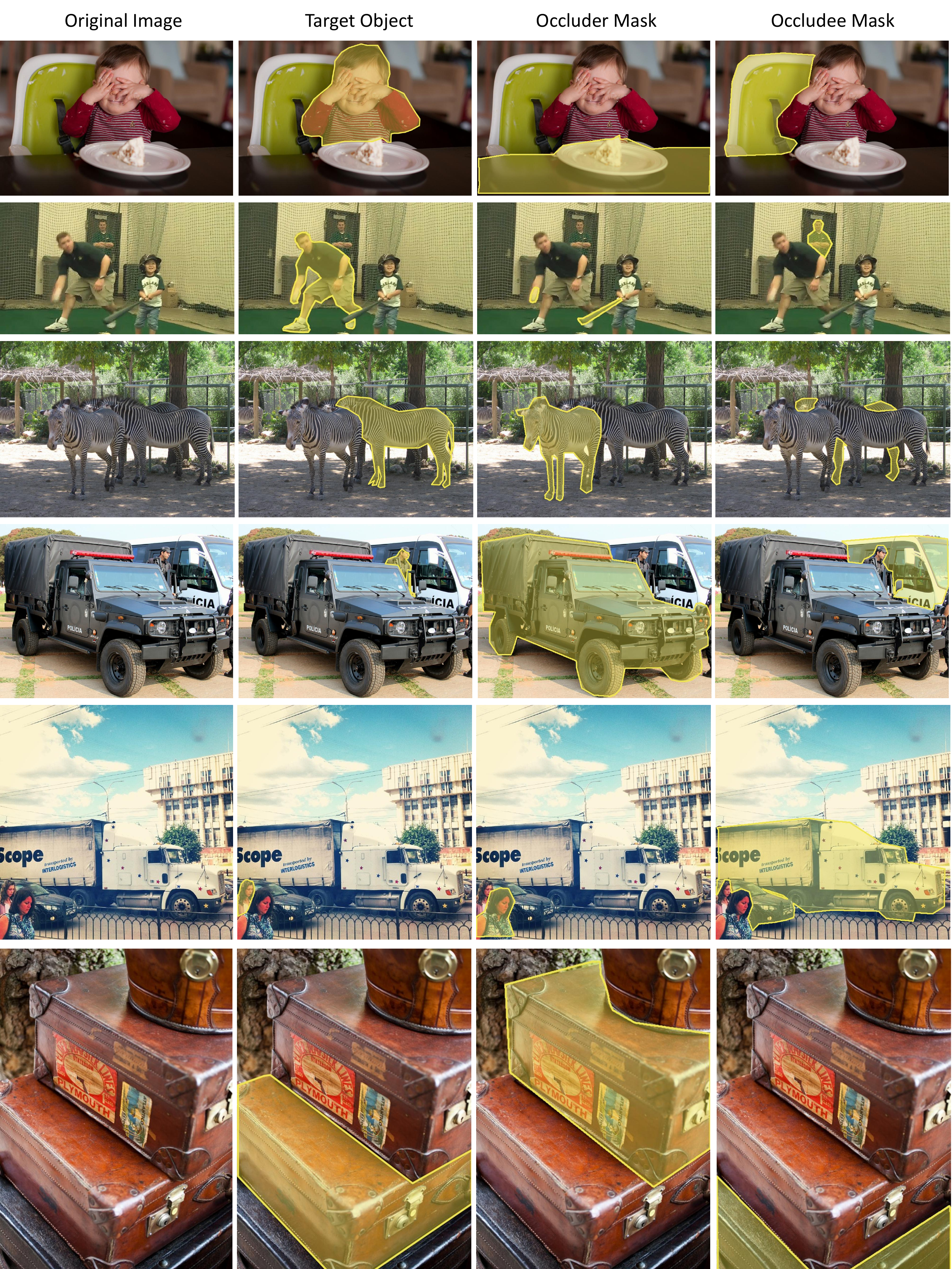}
		\caption{\color{bmvc_blue} \textbf{More examples of automatically generated training data.} For each row, from left to right is: original image, the object of interest (target object), its occluder mask, and its occludee mask.} 
  \label{figure:sup_generated_training_data}
 		\vspace{-1.5mm}
\end{figure*}

In Figure~\ref{figure:sup_generated_training_data}, 
we show visualisation of the results from our automatic pipeline that can infer
 occluder and/or occludee masks for 
target objects in COCO2017 train.

\clearpage
\section{Additional Qualitative Comparison on COCO}
\label{sec:sup_additional_qualitative_comparison}
\begin{figure*}[h!]
		\centering
		\includegraphics[trim=0.5cm 0cm 0.5cm 0cm, 
  width=0.8\linewidth
  ]{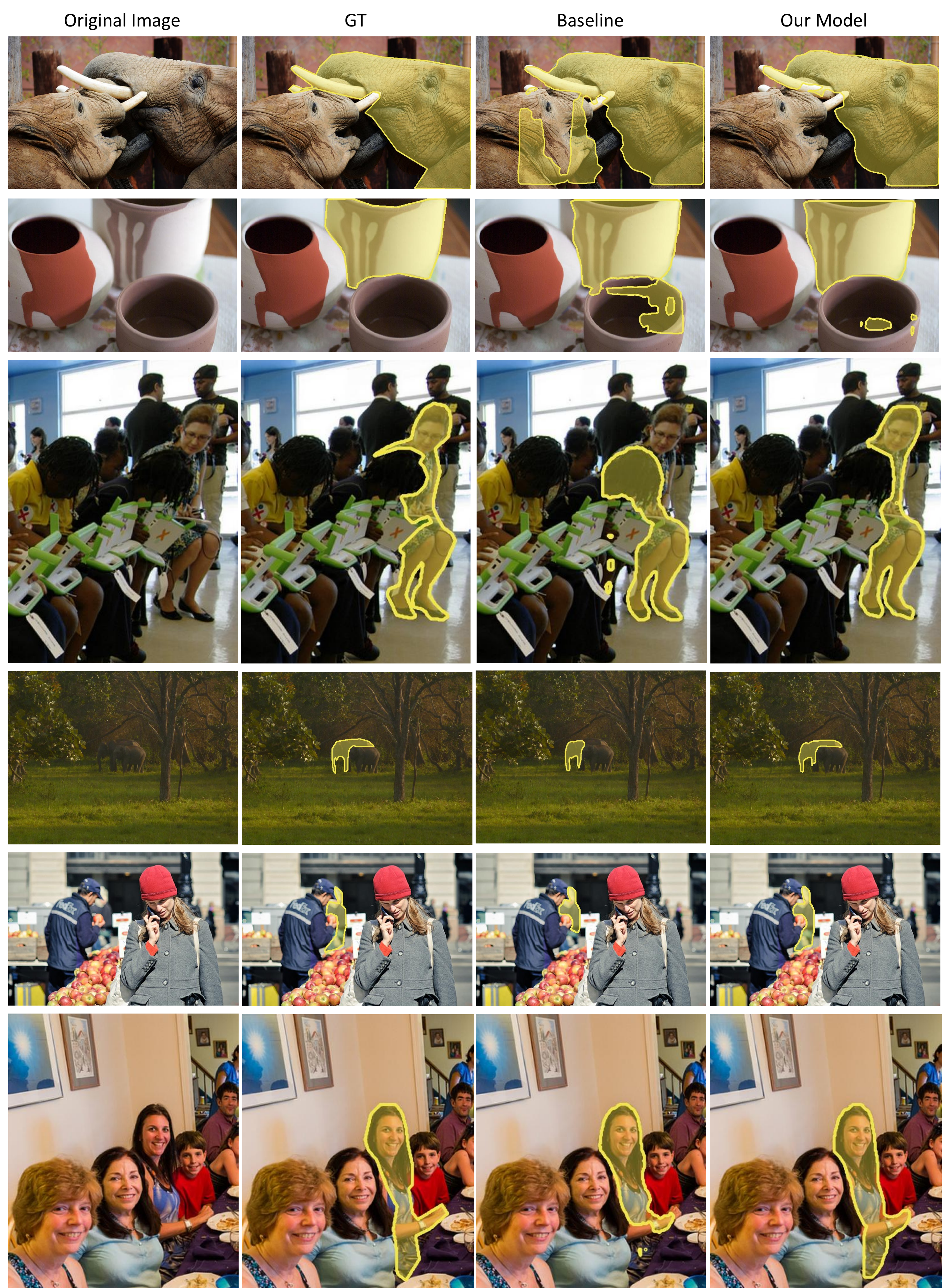}
		\caption{\color{bmvc_blue} \textbf{Additional qualitative comparison of our model with baseline on Occluded COCO.} Our model can systematically solve the common failure patterns of over-segmentation (the mask is too large, and includes part of occluder) (Row 1-3) and under-segmentation (the mask is too small) (Row 4-6) for partially occluded objects. 
  }
\label{figure:sup_additional_qualitative_comparison}
 		\vspace{-1.5mm}
\end{figure*}

\begin{figure*}[h!]
		\centering
		\includegraphics[trim=0.5cm 0cm 0.5cm 0cm,
  width=0.8\linewidth
  ]{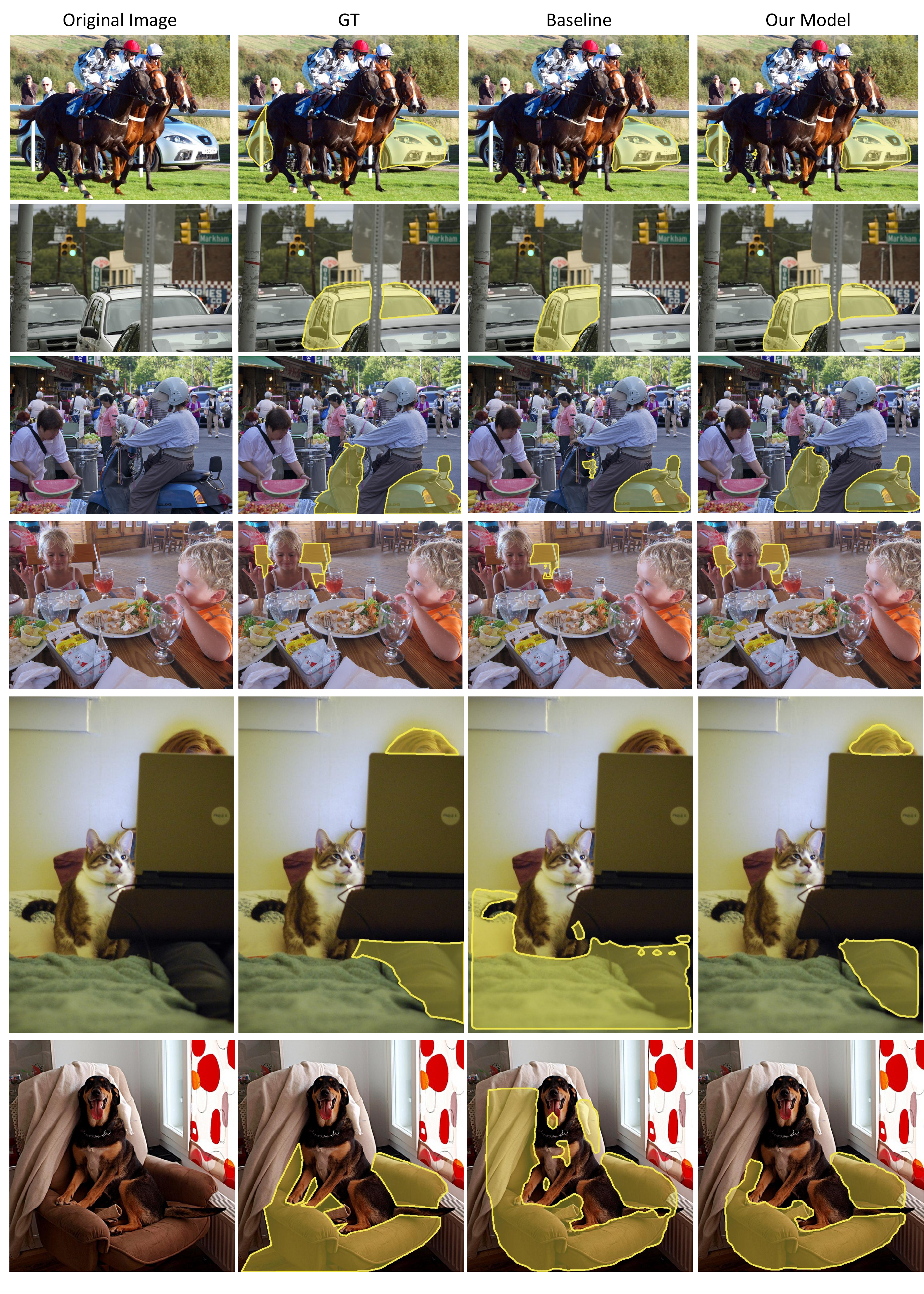}
		\caption{\color{bmvc_blue} \textbf{Additional qualitative comparison of our model with baseline on Separated COCO.} Our model could systematically solve the failure patterns of under-segmentation (the mask is too small, and only includes part of the separated object) (Row 1-4) and over-segmentation (the mask is too large, and also includes part of the occluder or surrounding objects) (Row 5-6) for separated objects. 
  }
\label{figure:sup_additional_qualitative_comparison_separated}
 		\vspace{-1.5mm}
\end{figure*}

Figures~\ref{figure:sup_additional_qualitative_comparison} and~\ref{figure:sup_additional_qualitative_comparison_separated} provide more qualitative detection examples to illustrate the effectiveness of our plugin over the baseline (Swin-T + Mask R-CNN) on partially occluded objects and separated objects. In both cases, our plugin can systematically solve two common failure patterns of the baseline detector: (1) Over segmentation, where part of the occluder or surrounding objects is included in the mask; (2) Under segmentation, where only part of the partially occluded / separated object is included in the mask.

\clearpage

\clearpage

\section{Results on Other Datasets}
\label{sec:sup_other_datasets}

In this section, we include additional experimental results for comparison between our model and baseline (Swin-T + Mask R-CNN) on OpenImages~\cite{openimages_2020}, OVIS~\cite{qi2022ovis} and KINS~\cite{qi2019kins}, as well as details for evaluation on each dataset. Section~\ref{sec:sup_openimages}, Section~\ref{sec:sup_ovis} and Section~\ref{sec:sup_kins} provide quantitative results and evaluation details for OpenImages, OVIS and KINS, respectively, while qualitative results are shown in Section~\ref{sec:sup_other_datasets_qualitative}.

\subsection{Quantitative Results on OpenImages}
\label{sec:sup_openimages}

OpenImages is a large-scale image dataset with manual annotations for some object masks together with the labels, indicating whether the object is occluded\slash truncated\slash crowd\slash depiction\slash inside or not. 
To evaluate the model's capability to detect partially occluded / separated objects, we collect a subset of objects with masks in OpenImages Test Set, 
which are labelled as occluded but not truncated/crowd/depiction/inside, 
and denote the subset as \textbf{Only Occluded OpenImages Test}. 
We further divide these objects into an Occluded set and a Separated set, depending on whether their masks are connected or not, 
like the division for COCO in the main paper. 
As a result, there are 3356 objects in total, with 2103 Occluded and 1253 Separated, in 2348 images. 

Table~\ref{tab:another_dataset_openimages} shows that our plugin can improve the performance of the baseline model in terms of recall and mIoU on both the Occluded objects and Separated objects. 
The recall and mIoU for Occluded objects only shows marginal improvement, because the selected “Occluded objects” are relatively easy and already well-detected by the baseline (over 75\%). Therefore, the advantage of our plugin on “Occluded objects” is not so significant.

\begin{table}[!htb]
\vspace{.2cm}
\centering
\tabcolsep=0.05cm
\begin{tabular}{ccccccccc}
\toprule
\multirow{2}*{Method}
&  \multicolumn{4}{c}{Only Occluded OpenImages Test} & \multicolumn{4}{c}{Sampled OVIS} \\ \cmidrule(lr){2-5}\cmidrule(lr){6-9} & \thead{Recall \\ Occluded} & \thead{Recall \\ Separated} &
\thead{mIoU \\ Occluded} & \thead{mIoU \\ Separated} &
\thead{Recall \\ Occluded} & \thead{Recall \\ Separated} &
\thead{mIoU \\ Occluded} & \thead{mIoU \\ Separated} \\ \midrule
Baseline & 1551 & 607 & 74.0 & 64.2 & 3960 & 1587 & \textbf{61.1} & 49.8\\ 
Baseline + Our Plugin &  \textbf{1569} & \textbf{672} & \textbf{74.1} & \textbf{65.4} & \textbf{3994} & \textbf{1673} & 60.5 & \textbf{50.3} \\ \bottomrule
\end{tabular}
\caption{\textcolor{bmvc_blue}{
Results on \textbf{Only Occluded OpenImages Test} and \textbf{Sampled OVIS}. For both datasets, the plugin slightly improves over the baseline's performance, particularly on Separated objects.}}
\label{tab:another_dataset_openimages}
\end{table}

\subsection{Quantitative Results on OVIS}
\label{sec:sup_ovis}

OVIS (Occluded Video Instance Segmentation) is a dataset with videos of occluded objects. For each object in the training set, 
it has mask annotations as well as a manual occlusion label to be ‘no occlusion’ / ‘slight occlusion’ / ‘severe occlusion’.
In order to evaluate on reasonably distinct frames, 
we pick 1 frame every 10 frames. 
Then we collect an evaluation image dataset (denoted as \textbf{Sampled OVIS}) containing 4443 images where we can calculate each detector’s recall of the Occluded objects and Separated objects (the division into 'Occluded' and 'Separated' is the same as in OpenImages). There are in total 7265 Occluded objects and 5187 Separated objects.

From Table~\ref{tab:another_dataset_openimages}, 
we can see that recall on Occluded objects and Separated objects of the baseline can be consistently boosted by the plugin. In terms of mIoU for Occluded/Separated objects, there is no significant improvement from the plugin.  These failure cases are mainly due to motion blur or require temporal context. We leave this to future work.

\subsection{Quantitative Results on KINS}
\label{sec:sup_kins}

As mentioned in Section~\ref{compare_stoa}, we evaluate on the KINS dataset
by directly evaluating the model that has been trained on COCO.
To handle the problem that KINS classes and COCO classes are different, we make a mapping from COCO classes as in Table~\ref{table:another_dataset_kins_class_mapping}. 
Note that ‘misc’ is not mapped to any COCO class, 
and ‘misc’ objects are not evaluated on.

\begin{table}[!htb]

\centering
\begin{tabular}{ccc}
\toprule
  KINS Class ID & KINS Class Name & Mapped to COCO Class Name  \\ \midrule
  1 & cyclist & person \\ 
  2 & pedestrain & person \\ 
  3 & rider & person \\ 
  4 & car & car \\ 
  5 & tram & train \\ 
  6 & truck & truck \\ 
  7 & van & truck \\ 
  8 & misc & none \\ \bottomrule
\end{tabular}
\vspace{.4cm}
\caption{\color{bmvc_blue} \textbf{Mapping from KINS classes to COCO classes} for evaluation.}
\label{table:another_dataset_kins_class_mapping}
\end{table}

Since ~\cite{yuan2021robust} has not released their code and model, 
it is difficult to make a fair comparison. 
In our evaluation, we test the baseline models and our model based on Swin-T + Mask R-CNN on the KINS test set, 
and calculate the mIoU following ~\cite{yuan2021robust}. 
In particular, during inference time, we input the GT box to the models because ~\cite{yuan2021robust} also inputs the GT amodal box to their model for instance segmentation inference which suits their setting. 

\begin{table}[!htb]
\centering
\begin{tabular}{cc}
\toprule
  Method & mIoU   \\ \midrule 
  Yuan \emph{et al}~\cite{yuan2021robust} & 67.2 \\ 
Swin-T + Mask R-CNN & 66.6 \\  
Swin-T + Mask R-CNN + Bi-Layer & 67.0 \\ 
Swin-T + Mask R-CNN + Our Plugin & \textbf{68.5} \\ \bottomrule
\end{tabular}
\vspace{.4cm}

\caption{\color{bmvc_blue} \textbf{Comparison with~\cite{yuan2021robust} on KINS}. With our plugin the baseline model can achieve a better performance than~\cite{yuan2021robust}.}
\label{table:sup_compare_alan}
\end{table}

Note that, the performance of our model is under-estimated with this evaluation protocol for the following reasons:
(i) KINS classes and COCO classes are different. We use a mapping from KINS classes to COCO classes, but this adds to the difficulty for our model to detect these KINS objects.
(ii) KINS annotations and COCO annotations are different, adding to the difficulty of adapting our models to test on KINS.
(iii) There could also be a domain shift.

Our model still outperforms the baseline model and the previous approach~\cite{yuan2021robust} (shown in Table~\ref{table:sup_compare_alan}), demonstrating the effectiveness of our plugin module.

\subsection{Qualitative Results on Other Datasets}
\label{sec:sup_other_datasets_qualitative}

\begin{figure*}[h!]
		\centering
		\includegraphics[trim=0.5cm 0cm 0.5cm 0cm,
  width=0.8\linewidth
  ]{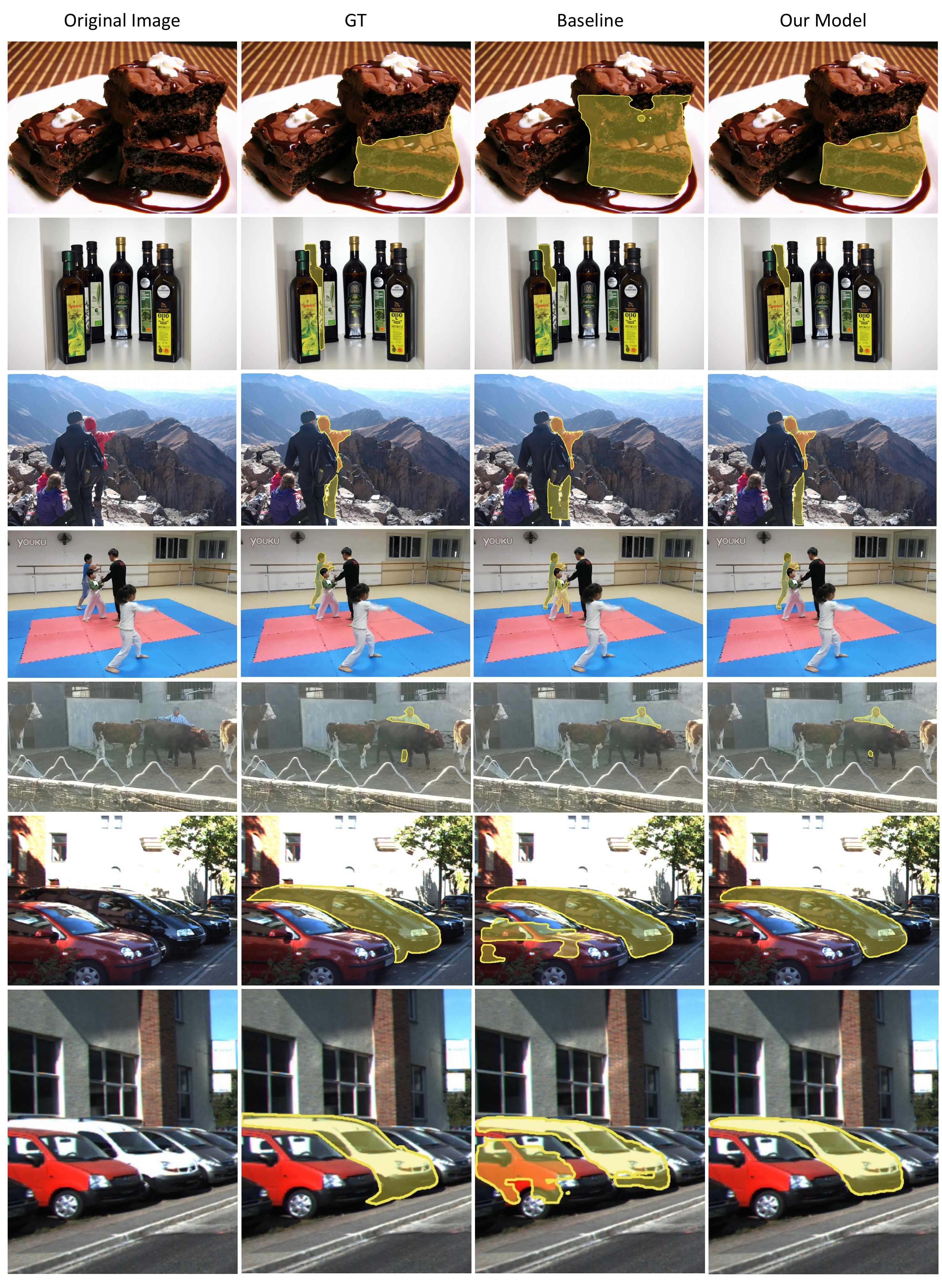}
		\caption{\color{bmvc_blue} \textbf{Qualitative comparison on other datasets.} Though not trained on these datasets, compared with the baseline, our model can solve the failure patterns of over-segmentation (including part of the occluder in the mask) (Row 1, 3, 4, 6, 7) and under-segmentation (Row 2, 5) for both partially occluded (Row 1, 2, 4, 6, 7) and separated objects (Row 3, 5). OpenImages: Row 1-3; OVIS: Row 4-5; KINS: Row 6-7. }
\label{figure:sup_other_datasets_qualitative_comparison}
 		\vspace{-1.5mm}
\end{figure*}

Figure~\ref{figure:sup_other_datasets_qualitative_comparison} shows examples where our plugin solves the baseline's failure patterns as mentioned in Section~\ref{sec:sup_additional_qualitative_comparison} on OpenImage, OVIS and KINS, qualitatively illustrating the effectiveness of our designed plugin when generalised to other datasets.

\clearpage

\section{Other Discussions}
\label{sec:sup_other_discussion}

\subsection{Number of Iterations}
\label{sec:sup_other_discussion_iteration}

For our current plugin, we apply two iterations of the tri-layer mask heads. We have also experimented with applying the module  three
times, and the comparison is shown in Table~\ref{table:sup_discussion_iteration}.

\begin{table}[!htb]

\centering
\begin{tabular}{ccccc}
\toprule
  Number of Iterations & Recall Occluded & Recall Separated & BBox mAP & Mask mAP  \\ \midrule
  Baseline* & 3264(58.81\%) & 1125(31.94\%) & 46.0 & 41.6 \\
  2 & \textbf{3434(61.87\%)} & \textbf{1208(34.30\%)} & 48.3 & 42.9 \\ 
  3 & 3401(61.28\%) & 1194(33.90\%) & \textbf{48.7} & \textbf{43.0}  \\ \bottomrule
\end{tabular}
\vspace{.4cm}
\caption{\color{bmvc_blue} \textbf{Comparison of different number of iterations on Swin-T + Mask R-CNN.} There is not much difference between 2 and 3 iterations. *Baseline denotes original Swin-T + Mask R-CNN without our plugin.}
\label{table:sup_discussion_iteration}
\end{table}

The mAP performance of three iterations is similar to using it twice, while
performance on occluded objects becomes worse.
For this reason, we only apply it twice.

\subsection{Class-Agnostic v.s. Class-Specific}
\label{sec:sup_other_discussion_agnostic}

For both bi-layer and tri-layer modelling, we have two choices of the occluder/occludee heads -- either to be class-agnostic or class-specific. Note that bi-layer modelling only has one extra occluder head to predict all surrounding objects of the target object, while tri-layer modelling has a pair of occluder/occludee heads to predict the occluders/occludees of the target object, and  can capture the occlusion ordering of different objects.

If an occluder/occludee head is class-agnostic, it only outputs one mask prediction; if the occluder/occludee head is class-specific, it outputs 80 mask predictions and the final result is the $i$-th mask prediction where $i$ is class prediction of the instance. The results of both bi-layer and tri-layer modelling under class-agnostic and class-specific settings are shown in Table~\ref{table:sup_discussion_agnostic}.

\begin{table}[h]

\centering
\tabcolsep=0.08cm
\begin{tabular}{cccccc}
\toprule
  \thead{Bi-Layer/Tri-Layer \\ Modelling} & \thead{Class-Specific/ \\ Class-Agnostic} & \thead{Recall \\ Occluded} & \thead{Recall \\ Separated} & BBox mAP & Mask mAP  \\ \midrule
  Baseline* & - & 3264(58.81\%) & 1125(31.94\%) & 46.0 & 41.6 \\
  Bi-Layer & Class-Specific & 3315(59.73\%) & 1147(32.57\%) & 46.3 & 42.0 \\ 
  Tri-Layer & Class-Specific & 3358(60.50\%) & 1166(33.11\%) & 46.2 & 42.2 \\ 
  Bi-Layer & Class-Agnostic & 3339(60.16\%) & 1147(32.57\%) & 46.3 & 42.2 \\ 
  Tri-Layer & Class-Agnostic & 3360(60.54\%) & 1159(32.91\%) & 46.3 & 42.2 \\ \bottomrule
\end{tabular}
\vspace{.4cm}
\caption{\color{bmvc_blue} \textbf{Comparison of bi-layer and tri-layer modelling under class-specific and class-agnostic settings.} In both settings, tri-layer modelling outperforms bi-layer modelling. *Baseline denotes original Swin-T + Mask R-CNN without our plugin.}
\label{table:sup_discussion_agnostic}
\end{table}

We can observe that in both class-agnostic and class-specific settings, tri-layer modelling outperforms bi-layer modelling.

\end{document}